\documentclass{article}

\usepackage{microtype}
\usepackage{graphicx}
\usepackage{makecell}
\usepackage{amssymb}
\usepackage{subfigure}
\usepackage{amsmath}
\usepackage{commath}
\usepackage{amsfonts}
\usepackage{booktabs} 
\usepackage{algorithm,algpseudocode}
\usepackage[dvipsnames]{xcolor}
\NewDocumentCommand{\code}{v}{%
    \texttt{\textcolor{OliveGreen}{#1}}%
}

\usepackage{hyperref}

\usepackage[accepted]{icml2024}

\usepackage{amsmath}
\usepackage{amssymb}
\usepackage{mathtools}
\usepackage{amsthm}

\usepackage[capitalize,noabbrev]{cleveref}

\theoremstyle{plain}

\theoremstyle{definition}

\theoremstyle{remark}

\usepackage[textsize=tiny]{todonotes}

\icmltitlerunning{}

\begin{document}

\twocolumn[
\icmltitle{Enhancing Stability for Large Language Models Training in Constrained Bandwidth Networks}




\begin{icmlauthorlist}
\icmlauthor{Yun Dai}{comp}
\icmlauthor{Tejas Dharamsi}{comp}
\icmlauthor{Pin-Lun (Byron) Hsu}{comp1}
\icmlauthor{Tao Song}{comp}
\icmlauthor{Hamed Firooz}{comp}
\end{icmlauthorlist}

\icmlaffiliation{comp}{Foundational AI Technologies (FAIT), LinkedIn}
\icmlaffiliation{comp1}{AI Platforms, LinkedIn}

\icmlcorrespondingauthor{Yun Dai}{yudai@linkedin.com}
\icmlcorrespondingauthor{Tejas Dharamsi}{tdharamsi@linkedin.com}
\icmlcorrespondingauthor{Pin-Lun (Byron) Hsu}{byhsu@linkedin.com}
\icmlcorrespondingauthor{Tao Song}{tsong@linkedin.com}
\icmlcorrespondingauthor{Hamed Firooz}{hfirooz@linkedin.com}

\icmlkeywords{Machine Learning, ICML}

\vskip 0.3in
]



\printAffiliationsAndNotice{}  

\begin{abstract}
Training extremely large language models (LLMs) with billions of parameters is a computationally intensive task that pushes the limits of current data-parallel training systems. While techniques like ZeRO++ \cite{wang2024zero} have enabled efficient distributed training of such giant models on inexpensive low-bandwidth clusters, they can suffer from convergence issues due to potential race conditions in the hierarchical partitioning (hpZ) scheme employed to reduce cross-machine communication. In this work, we first show how these race conditions cause instability when training models with billions of parameters. We then propose a modification to the partitioning algorithm that addresses these convergence challenges while maintaining competitive training efficiency. Empirical evaluation on training the multi-billion parameters Falcon Models and LLama-2 models demonstrates the updated algorithm's ability to achieve reliable convergence on these massive models, where stock ZeRO++ \textit{hpZ} fails to converge. The updated algorithm enables robust training of larger models with 98\% throughput and model training speed improvement without sacrificing the quality of convergence. 


\end{abstract}

\section{Introduction}
\label{introduction}
 The ambition to create more expansive and capable AI models has catalyzed a continual push to achieve greater scale in machine learning model training frameworks. Transformer-based model architectures like GPT \cite{achiam2023gpt}, Falcon \cite{almazrouei2023falcon}, Mistral \cite{jiang2023mistral} LLaMA \cite{touvron2023llama} and many other have surpassed billions of parameters, enabling superior performance on a wide range of language understanding and generation tasks. The remarkable capabilities unlocked by these massive models have spurred surging interest from both industry and academia to explore the limits of scale for large AI models. However, the immense computational requirements for training these giant neural networks stretch the limits of modern hardware and distributed training frameworks. Crucially, the pursuit of ever-larger models risks exacerbating disparities, as the specialized acceleration hardware required remains cost-prohibitive and inaccessible for many.  Democratizing access to train massive models efficiently on modest, commodity hardware is vital to ensure equitable AI development globally.

Data parallelism, where a model's parameters are partitioned across multiple accelerator devices (e.g. CPUs, GPUs, TPUs) during training, has been a key enabling technique for training large deep neural network models. Classic data-parallel implementations like BytePS \cite{258953}, Horovod \cite{sergeev2018horovod} and PyTorch DDP \cite{10.14778/3415478.3415530} rely on replicating the full model across all devices, leading to substantial memory overheads that impose a hard limit on maximum model size. 

Megatron-LM \cite{shoeybi2020megatronlm} from NVIDIA attempted to scale by leveraging model parallelism in addition to data parallelism. However, this approach required specialized high-bandwidth interconnects like NVLink and InfiniBand that are not widely accessible to most developers and researchers.

In contrast, methods like ZeRO \cite{rajbhandari2020zero} took a different approach - eliminating redundant parameter copies across data-parallel devices through intelligent parameter partitioning schemes. ZeRO-Offload \cite{ren2021zero} further optimized memory usage by offloading activations and optimizer states to CPU memory during the respective forward and backward passes.  While more memory-efficient, these ZeRO techniques still relied heavily on frequent inter-node communication for collective operations. 

A key bottleneck faced by both Megatron-LM and ZeRO was the relatively low inter-node communication bandwidth compared to the intra-node bandwidth within a single multi-accelerator system. As these approaches scaled out to larger node counts for training trillion-parameter models, the inter-node communication overheads became pronounced. This communication bottleneck imposed severe scaling limits when using commodity multi-node clusters without ultra high-speed networking fabrics \cite{liang2024communicationefficient}.

To enable large scale training in network constraint environment, ZeRO++ algorithm \cite{wang2024zero} introduced a set of efficient parallelization strategy leveraging Quantized Weight Communication for ZeRO (\textit{qwZ}), Quantized Gradient Communication for ZeRO (\textit{qgZ}), Hierarchical Weight Partition for ZeRO (\textit{hpZ}) hierarchical partitioning to minimize cross device communication volume. ZeRO++ has been pivotal, enabling training of unprecedentedly massive models like GPT-3 \cite{brown2020language} with 175B parameters and Turing NLG 530B \cite{smith2022using} on relatively small GPU clusters.

Despite its scalability claims, the hierarchical partitioning algorithm in the original ZeRO++ implementation can encounter convergence issues on extremely large models like Falcon (40B) and Meta's LLaMA (70B) during full-parameter fine-tuning, depending on the hardware and fabric setup. These issues lead to unreliable and unstable training runs. The convergence failures stem from a subtle race condition between the asynchronous parameter partitioning operation and collective communication primitives, such as \code{AllGather}, in ZeRO++'s algorithm. Incorrect ordering can lead to corrupted parameter values being communicated across devices, causing training instability and divergence. Addressing these convergence pitfalls is crucial to fully unleash the potential of giant language model training in a reliable and robust manner over commodity hardware.

In this work, we perform an in-depth analysis of the root causes behind ZeRO++ \textit{hpZ}'s convergence failures on publicly available large language models like Falcon and Llama. We identify the key synchronization bugs in ZeRO++'s hierarchical partitioning scheme that trigger these failures. Based on our findings, we propose a simple yet effective modification to ZeRO++ that introduces explicit CUDA synchronization points to ensure all parameter partitioning completes correctly before any collective communication over the partitioned data. Updated algorithm restores reliable convergence for training giant transformer models using ZeRO++ without impacting its highly coveted computational efficiency and scalability advantages.

\section{Background}

As we push to hundreds of billions parameters models, the memory becomes a burden for training. The ZeRO3 algorithm shards model parameters, gradients, and optimizer states to significantly reduce the memory footprint. However, this reduction comes at the cost of increased communication overhead. Specifically, the algorithm requires \code{AllGather} operations on the weights during both the forward pass to compute activations and the backward pass to compute gradients, followed by reduce-scatter operations to distribute the gradients across accelerators.

This results in a high dependence on the communication bandwidth within the cluster. Recent research \cite{liang2024communicationefficient} \cite{li2019evaluating} \cite{ren2019performance} has shown that the latency for inter-node GPU communication is typically higher compared to intra-node GPU communication. This can be attributed to the differences in interconnect technologies such as NVLink NVSwitch~\cite{nvidia2024nvlink} for intra-node and Infiniband~\cite{nvidia2024infiniband} for inter-node communication. As a result, the training process often encounters bottlenecks due to slower inter-node communication, making the inter-node network more likely to become the system bottleneck. Improving inter-node network speed can lead to significant performance gains for multi-GPU applications.

To address these bottlenecks, several approaches have been proposed to maximize intra-node communication while minimizing inter-node communication. For example, PyTorch's hybrid shard FSDP~\cite{zhao2023pytorch} performs ZeRO3 only within a node and uses \code{AllReduce} to synchronize gradients across nodes, similar to the Distributed Data Parallel approach. Another approach, MiCS~\cite{zhang2022mics}, creates subgroups of GPUs with high intra-group bandwidth, limiting \code{AllGather} operations to within these subgroups and using AllReduce for inter-group communication to minimize the inter-subgroup communication overhead.

ZeRO++~\cite{wang2024zero} extends these ideas with the introduction of hierarchical partitioning \textit{hpZ}, quantized gradient \textit{qgZ}, and quantized weight \textit{qwZ} schemes. The \textit{hpZ} scheme ensures a full copy of the model within a node and uses \code{AllReduce} to synchronize gradients across nodes, effectively reducing inter-node communication. The \textit{qgZ} scheme quantizes gradients before \code{ReduceScatter} operations, and the qwZ scheme quantizes weights before \code{AllGather} operations, both of which further optimize communication efficiency.

\section{Algorithm}\label{sec:algorithm}
\begin{figure}
    \centering
    \includegraphics[width=1\linewidth]{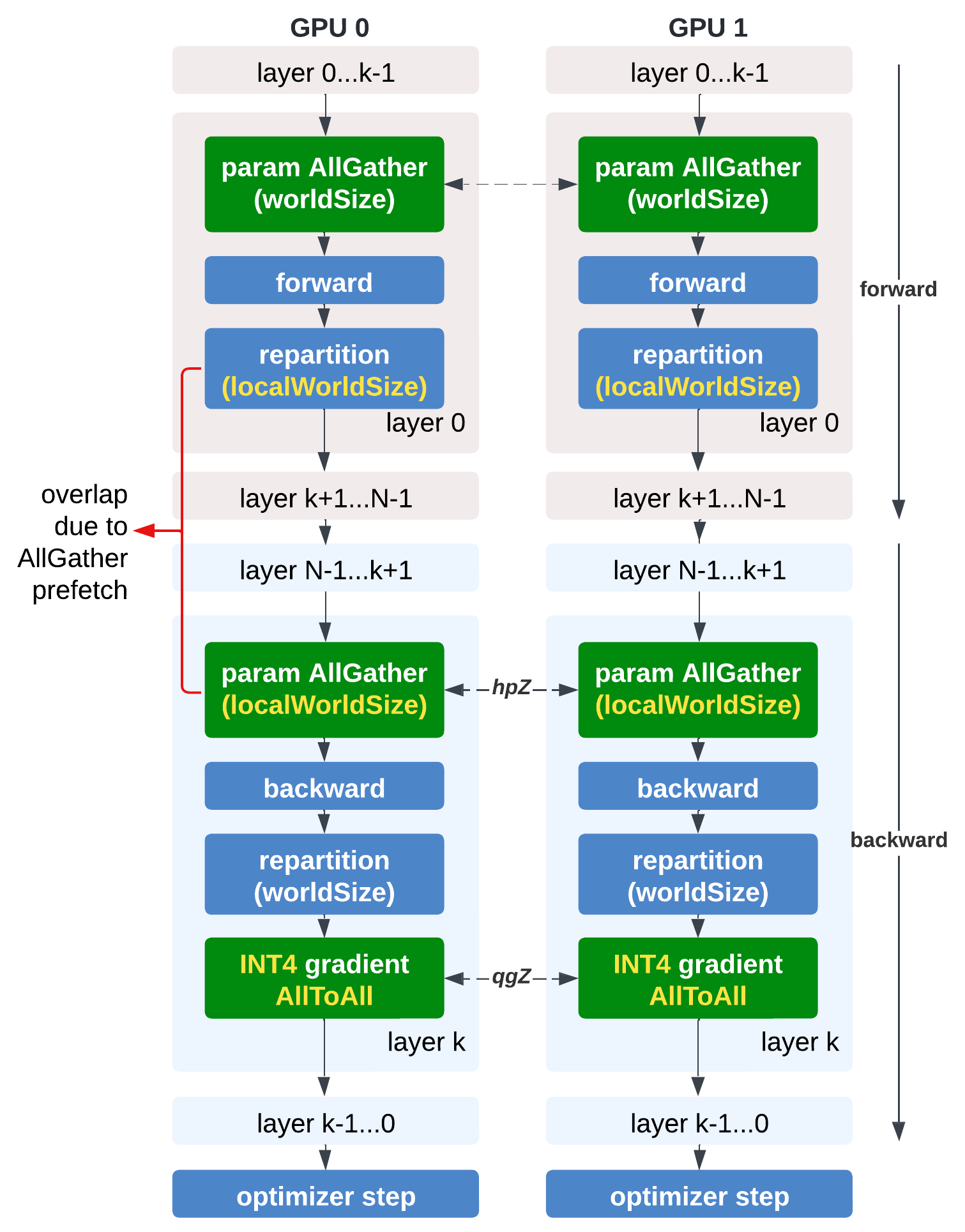}
    \caption{An end-to-end training step on a model with $N$ layers with ZeRO++ $qgZ$ and $hpZ$. Forward and backward pass on layer $k$ is expanded. With ZeRO3 prefetch, the consequent \texttt{AllGather} kernel for the backward pass can be immediately enqueued while post-forward repartition is still in progress.}
    \label{fig:zpp-algo}
\end{figure}

ZeRO++ \textit{hpZ} reduces communication overhead by eliminating cross-node \code{AllGather} communication during the backward pass, with an extra cost of memory. After the forward pass of a layer is done, instead of re-spreading its full weights across all GPUs, ZeRO++ \textit{hpZ} partitions the weights into a secondary copy which is replicated on each node. \code{AllGather} in the backward pass thus operates on the secondary copy and will only involve intra-node communication, which has multiple factors higher bandwidth than inter-node. Figure \ref{fig:zpp-algo} provides an illustration of a full training step with \textit{qgZ} and \textit{hpZ} enabled.

\begin{algorithm}[H]
\caption{ZeRO++ \textit{hpZ} with prefetch}\label{alg:cap}
\begin{algorithmic}[1]
\algnewcommand{\LineComment}[1]{\State \(\triangleright\) #1}
\Require
\Statex $worldSize: P$
\Statex $secondaryWorldSize: P'$
\Statex $model: \mathcal{M} \& = \{ L_1, L_2, \ldots, L_N \}$

\Function{PrefetchAllGather}{}
   \For{$L_k \in \text{next modules to be executed}$}
      \If{is\_forward($L_k$)}
         \State Enqueue AllGather($L_k$, $P$)
      \Else
      \color{blue}
         \Repeat
            \State wait
         \Until{MemcpyD2D on $L_{k,second}$ finishes}
         \color{black}
         \State \text{Enqueue AllGather}($L_k$, $P'$)
      \EndIf
   \EndFor
\EndFunction
\While{model not converged}
    \LineComment{Forward pass}
    \For{$i = 1, 2, ..., N$}
        \State Ensure AllGather($L_i$, $P$) finished
        \State PrefetchAllGather()
        \State $L_i$.forward()
        \State $L_{i,second} \gets empty(\frac{|L_i|}{P'})$
        \State Copy to $L_{i,second}$ \Comment{Async MemcpyD2D}
    \EndFor
    \Statex \ 
    \LineComment{Backward pass}
    \For{$i = N, N-1, ..., 1$}
        \State Ensure AllGather($L_i$, $P'$) finished
        \State PrefetchAllGather()
        \State $L_i$.backward()
        \State repartition($P$)
        \LineComment{Replaced with INT4 AllToAll if with $qgZ$}
        \State ReduceScatter($\nabla L_i$, $P$)
    \EndFor
    \State optimizer.step()
\EndWhile
\end{algorithmic}
\end{algorithm}

During partition, the secondary copy tensor gets allocated as \code{torch.empty} of the following shape

\begin{equation}
 \|L_{i, second} \| \leftarrow \frac{N}{\textit{secondaryWorldSize}}
\end{equation}

where $N$ is the total number of elements in the weights and $secondaryWorldSize$ is typically equal to the number of GPUs per node. The weights corresponding to the local rank then get copied from the full parameter tensor to the secondary copy.

However, since the Memcpy is from and to tensors both allocated on GPU, hence an asynchronous D2D (device-to-device) copy, there is no guarantee that the secondary copy is settled when the following \code{AllGather} kernel on it is launched.

With prefetch in ZeRO, which enqueues \code{AllGather} kernels for following modules beforehand instead of as late as when the backward pass actually happens, a race condition may happen: at the time when a module is still being partitioned into the secondary copy, the \code{AllGather} kernel for the backward pass on it can be immediately enqueued and launched. This results in \code{AllGather} aggregating on arbitrarily initialized tensor values, leading to model instability during training and often results in Not-a-Number ($NaN$) in aggregated parameter values and hence the observed loss.

The race condition is mainly the result of full asynchronization between two operators, Memcpy and \code{AllGather}. To avoid such a race condition, we add a CUDA synchronization operator between D2D Memcpy and \code{AllGather}. The sync operator first makes sure that the Memcpy is finished and upon success of Memcpy D2D, it enqueues the \code{AllGather} and therefore removes the race condition between the two. The modified algorithm is illustrated in Algorithm \ref{alg:cap}.

In next section, we show that due to the asynchronous property of Memcopy, there is no guarantee that the secondary copy is settled before the \code{AllGather}, and as a result, the model performance is not predictable. This is particularly evident for large models trained on commodity bandwidth-limited networks.

\section{Experimentation}
\subsection{Experimentation setup}
All experiments are conducted on NVIDIA-A100 GPUs. Eight GPUs compose one GPU node, where GPUs within a node are connected using NVIDIA NVLINK with 600 GB/s bandwidth. To simulate commodity low bandwidth connections, we use 1$\times$ 12.5Gbps Ethernet NIC per node.

We leverage two families of off-the-shelf publicly available large language models for our experiments with different sizes: 1) Llama-2 \cite{touvron2023llama}, 2) Falcon \cite{almazrouei2023falcon}. We perform full parameter fine-tuning using the MMLU dataset \cite{hendrycks2020measuring} on these models.

Here we define training as unstable if the loss during training diverges to $NaN$ or the training loss does not decrease with the same hyperparameters. We measure the throughput of the training by determining how many tokens are processed in one second given a full GPU node, a.k.a tokens/s/node.

\subsection{Divergence Analysis}
To demonstrate model training instability, we train various off-the-shelf LLMs with and without stock ZeRO++ \textit{hpZ} and compare the stability with modified \textit{hpZ} outlined in Algorithm \ref{alg:cap}. As demonstrated in Table \ref{table:stability}, for many publicly available large models, the training is unstable with the original ZeRO++ hierarchical partitioning.

\begin{table}[h!]
\centering
\begin{tabular}{l||c|c|c}
Model & without \textit{hpZ} & with \textit{hpZ} & \makecell{modified \\ \textit{hpZ}} \\
\specialrule{1.5pt}{0pt}{0pt}
\hline
Llama-2-7B & $\checkmark$ & $\times$ & $\checkmark$ \\
\hline
Llama-2-13B & $\checkmark$ & $\times$ & $\checkmark$ \\
\hline
Llama-2-70B & $\checkmark$ & $\times$ & $\checkmark$ \\
\hline
Falcon-40B & $\checkmark$ & $\times$ & $\checkmark$ \\
\hline
\end{tabular}
\caption{Stock ZeRO++ causes instability in model training due to the race condition between AllGather and Memcopy.}
\label{table:stability}
\end{table}

The modified algorithm adds a synchronization operation to guarantee the healthiness of \code{AllGather}. This impacts the throughput of model training. To measure such impact, we fine-tune multiple off-the-shelf models on the MMLU dataset and measure the throughput of the training.

\begin{table}[h!]
\centering
\begin{tabular}{l||c|c|c}
Model & \textit{qgZ} only & \textit{qgZ} + \textit{hpZ} & \makecell{\textit{qgZ} + \\ modified \textit{hpZ}} \\
\specialrule{1.5pt}{0pt}{0pt}
\hline
Llama-2-7B & 2880 & 5065.4 & 5013 (+74\%) \\
\hline
Llama-2-13B & 2101 & 2510.5 & 2521.4 (+20\%) \\
\hline
Llama-2-70B & 358 & 488.1 & 456.9 (+27\%) \\
\hline
Falcon-40B & 444.5 & 890 & 881.9 (+98\%) \\
\hline
\end{tabular}
\caption{Modified \textit{hpZ} has a small impact on training throughput compared to stock ZeRO++. \textit{qgZ} is enabled for all runs as control. The numbers in paranthesis shows the speed up as compares to baseline of \textit{qgZ} only}
\label{table:throuput}
\end{table}

As shown in Table \ref{table:throuput}, the async operation between \code{AllGather} and memory copy in ZeRO++ sometimes results in lower training throughput compared to stock \textit{hpZ}, but as explained in Table \ref{table:stability}, it comes with instability and divergence costs. The modified hierarchical partitioning algorithms have up to 98\% higher throughput compared to when \textit{hpZ} is not enabled while keeping the training stability unchanged.

\vspace{-5mm}
\begin{figure}[h]
\centering
\includegraphics[width=1\linewidth]{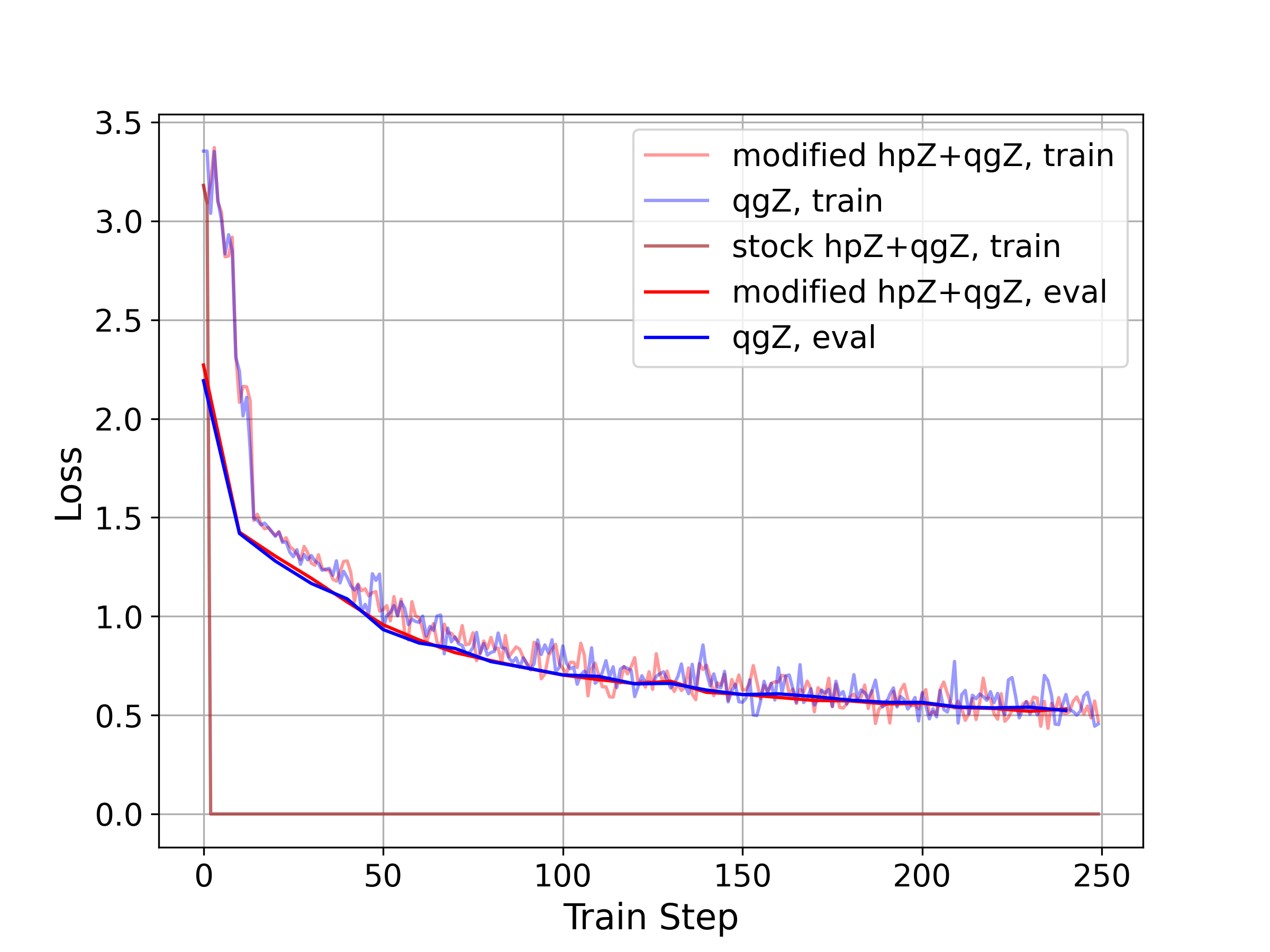}
\caption{Validation loss convergence per optimization step without \textit{hpZ} and with modified \textit{hpZ} from Algorithm \ref{alg:cap}, tested on Llama-2-7b for MMLU dataset. Brown curve demonstrates convergence issue in training without fix.}
\label{fig:val_loss}
\end{figure}

Despite having better training throughput, the modified hpZ has no impact on model optimization performance and the training is stable. Figure \ref{fig:val_loss} shows the validation loss during training for MMLU without \textit{hpZ} and with modified \textit{hpZ} for Llama-2-7B. As one can see, there is almost no difference in validation loss convergence per optimization step while from Table \ref{table:throuput} the modified \textit{hpZ} converges faster in wall-clock time.

\section{Conclusion}

In this work we identified and addressed a key convergence issue that affects the training of large language models using the ZeRO++ algorithm on commodity hardware with limited network bandwidth. We analyzed the root cause of the instability, which stems from a race condition between the asynchronous parameter partitioning and collective communication operations in ZeRO++'s hierarchical partitioning scheme. We resolve this bottleneck by introducing explicit CUDA synchronization. This ensures parameter partitioning completes correctly before any collective communication over the partitioned data occurs. 
Our empirical evaluation demonstrated that the updated algorithm restores reliable convergence when training giant transformer models like the 40B parameter Falcon and 70B parameter LLama-2 on the challenging MMLU dataset.



\nocite{langley00}

\bibliography{example_paper}
\bibliographystyle{icml2024}

\end{document}